\documentclass[letterpaper,10 pt,conference]{ieeeconf}

\IEEEoverridecommandlockouts                              

\usepackage{graphicx}
\usepackage{amsmath,amssymb}
\usepackage{mathtools,bm}

\newtheorem{proposition}{Proposition}
\usepackage{pifont}
\usepackage{makecell}
\usepackage{multirow}

\usepackage[labelfont=bf,margin=0.25in,font=small]{caption}
\usepackage{subcaption}
\usepackage{wrapfig}
\usepackage{enumerate}

\usepackage{array}
\usepackage[table,dvipsnames]{xcolor}
\usepackage{makecell, cellspace}

\usepackage{xargs} 

\usepackage{hyperref} 
\usepackage{algorithm}
\usepackage{algpseudocode}
\usepackage{booktabs}
\usepackage{cite}
\definecolor{lightblue}{rgb}{0.85,0.91,0.95}

\title{\LARGE \bf
Unified Generation-Refinement Planning: Bridging Guided Flow Matching and Sampling-Based MPC for Social Navigation
}

\author{
  Kazuki Mizuta, Karen Leung\\
  University of Washington, Seattle, WA\\
  \texttt{\{mizuta,kymleung\}@uw.edu} \\
}
\author{Kazuki Mizuta$^{1}$ and Karen Leung$^{1,2}$
\thanks{Kazuki Mizuta is partially supported by the Nakajima Foundation.}
\thanks{This work was supported by the National Science Foundation award under Grant No. 2430686 and 2440861.}
\thanks{$^1$University of Washington, Department of Aeronautics and Astronautics, $^2$ NVIDIA,
        Contact: {\tt\small \{mizuta, kymleung\}@uw.edu }}
}

\begin{document}
\maketitle
\thispagestyle{empty}
\pagestyle{empty}

\begin{abstract}
    Robust robot planning in dynamic, human-centric environments remains challenging due to multimodal uncertainty, the need for real-time adaptation, and safety requirements. Optimization-based planners enable explicit constraint handling but can be sensitive to initialization and struggle in dynamic settings. Learning-based planners capture multimodal solution spaces more naturally, but often lack reliable constraint satisfaction. In this paper, we introduce a unified generation-refinement framework that combines \textit{reward-guided conditional flow matching} (CFM) with model predictive path integral (MPPI) control. Our key idea is a \textit{bidirectional information exchange} between generation and optimization: reward-guided CFM produces diverse, informed trajectory priors for MPPI refinement, while the optimized MPPI trajectory warm-starts the next CFM generation step. Using autonomous social navigation as a motivating application, we demonstrate that the proposed approach improves the trade-off between safety, task performance, and computation time, while adapting to dynamic environments in real-time. The source code is publicly available at \url{https://cfm-mppi.github.io}.
\end{abstract}

\section{Introduction} \label{sec:introduction}

Robust planning in dynamic human-centric environments (e.g., warehouses, sidewalks, airports) requires fast replanning, safety constraint enforcement, and reasoning over uncertain, multimodal human behavior \cite{RudenkoPalmieriEtAl2020,LeungSchmerlingEtAl2020,SchmerlingLeungEtAl2018,NishimuraIvanovicEtAl2020}. 
Optimization- and learning-based methods have shown strong performance in settings like human-robot interaction \cite{NishimuraIvanovicEtAl2020,EverettChenEtAl2018}, mobile autonomy \cite{WilliamsDrewsEtAl2016,ScharfAcikmeseEtAl2017}, and manipulation \cite{ChiFengEtAl2023}, but struggle to meet all demands simultaneously.
Optimization-based methods \cite{AcikmeseBlackmore2011} can provide explicit constraint handling and strong online performance, but often depend on good initialization and may struggle in dynamic, uncertain, or highly nonconvex environments \cite{GrootBritoEtAl2021}.
In contrast, learning-based approaches (e.g., diffusion policies \cite{SohlDicksteinWeissEtAl2015,HoJainEtAl2020,JannerDuEtAl2022}) are well-suited to modeling multimodal solution spaces, but they do not explicitly enforce safety constraints and can be computationally expensive at inference time.

Conditional flow matching (CFM) \cite{LipmanChenEtAl2023} has emerged as a promising generative framework that achieves generation quality comparable to diffusion models while requiring fewer inference steps.
Recent works have begun to apply CFM to robotics problems \cite{ChisariHeppertEtAl2024,YeGombolay2024}, but its integration with safety-critical online planning in dynamic environments remains relatively unexplored.

\begin{figure}[t]
  \centering
  \includegraphics[width=\linewidth]{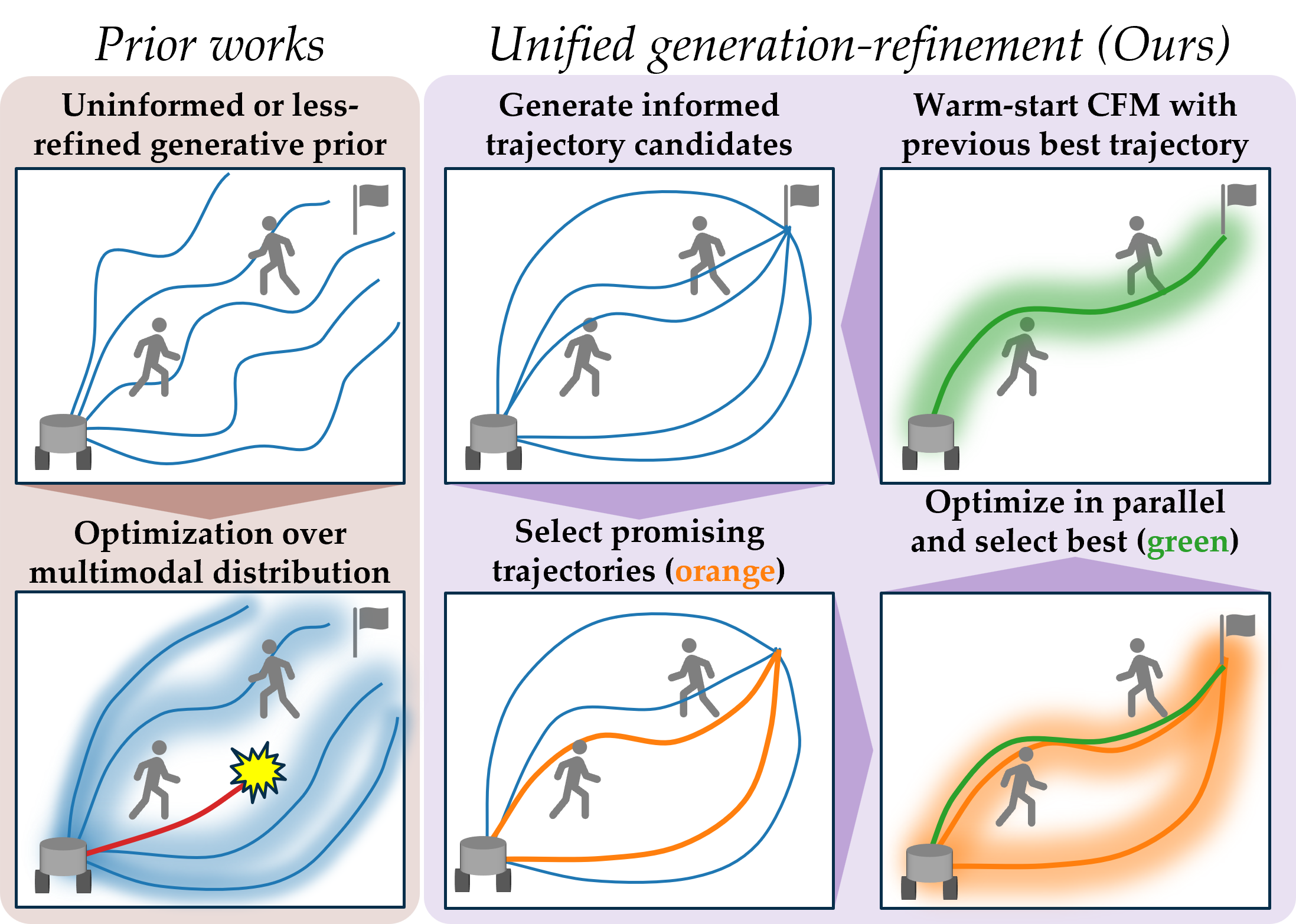}
  \captionsetup{width=\linewidth}
  \caption{Overview of the proposed unified planning framework for dynamic environments: At each planning step, our conditional flow matching (CFM) model generates context-aware and multimodal trajectory candidates guided by a reward function.
  Promising candidates are selected, then refined, and the best trajectory is selected and executed.
  The optimal trajectory warm-starts the CFM generation for the next planning step.
  }
  \label{fig:hero}
  \vspace{-5mm}
\end{figure}

To address this, we propose a unified generation-refinement planning framework that combines reward-guided CFM with online optimization-based planning, specifically model predictive path integral (MPPI) control \cite{WilliamsDrewsEtAl2016}.
As illustrated in Fig.~\ref{fig:hero}, reward-guided CFM generates diverse, context-aware trajectory priors for MPPI refinement, while the optimized MPPI trajectory warm-starts the next CFM generation step, forming a bidirectional feedback loop.

A key challenge is preserving multimodality during refinement, since naively aggregating distinct trajectory modes in a standard MPPI update can collapse them into a poor intermediate solution. 
We therefore introduce a mode-selective refinement strategy that selects promising CFM trajectories, refines each independently with MPPI in parallel, and executes the best refined trajectory.

We evaluate the method on autonomous social navigation, where a robot must safely move through crowds of pedestrians with uncertain and multimodal motion patterns. 
Our results show improved trade-offs between safety, task performance, and computation time compared with standalone baselines.

\noindent\textbf{Statement of contributions.} Our contributions are fourfold:
\textbf{(i)} A unified generation-refinement planning framework that bidirectionally couples reward-guided CFM and MPPI, combining generative multimodality with online optimization.
\textbf{(ii)} A reward-guided CFM sampling mechanism that adapts trajectory generation to new planning objectives at test time without retraining.
\textbf{(iii)} A mode-selective MPPI strategy that refines multimodal trajectory priors without collapsing distinct modes into suboptimal solutions.
\textbf{(iv)} Evaluation on social navigation tasks, demonstrating improved trade-offs between solution quality, safety, and generation time over standalone approaches.

\section{Related Work} \label{sec:related_work}
This section reviews relevant prior work in optimization-based trajectory planning approaches and recent advances in data-driven generative models for motion planning.

\noindent\textbf{Learned initializations for optimization-based trajectory planning.} 
Optimization-based methods are attractive for safety-critical trajectory planning because they can explicitly handle constraints and offer strong online performance. 
However, they typically rely on accurate system models and good initializations to converge to high-quality solutions. 
Common initialization strategies, such as straight-line trajectories or previous solutions, are simple but usually unimodal, which limits performance in highly nonconvex settings such as dynamic obstacle avoidance.

Recent work has addressed this limitation by learning informed initializations or sampling distributions conditioned on the environment \cite{SacksBoots2023,BanerjeeLewEtAl2020,GuffantiGammelliEtAl2024,HondaAkaiEtAl2023,PowerBerenson2024,LiBeeson2025,SharonyYangEtAl2025}. 
These approaches can substantially improve optimization, but they often rely on strong supervision such as expert or near-optimal demonstrations, and may be difficult to scale to dynamic multi-agent settings or to transfer across environments.

Other recent approaches improve online responsiveness by biasing sampling distributions with explicitly computed backup policies (e.g., contingency plans) \cite{TrevisanAlonsoMora2024,JungEstornellEtAl2025}. 
While effective, these methods require solving additional planning problems, incorporating system-specific knowledge, or relying on handcrafted heuristics, which adds computational overhead and can reduce generality.
In contrast, our approach learns a sampling distribution from real-world data and biases it online through a lightweight reward-guidance mechanism without requiring additional training.

\noindent\textbf{Diffusion policies.}
Deep generative models have shown strong performance in learning complex distributions \cite{KothariKreissEtAl2022,NieVahdatEtAl2021}.
Guided diffusion models \cite{SohlDicksteinWeissEtAl2015,HoJainEtAl2020,HoSalimans2021} are particularly attractive because of their expressiveness, stable training, and ability to incorporate reward-based guidance, which has made them effective for planning and trajectory generation \cite{ChiFengEtAl2023,JannerDuEtAl2022,CarvalhoLeEtAl2023,MengFan2024,ZhongRempeEtAl2023,XiaoWangEtAl2025}.
However, they face two key limitations in real-time planning: (i) high computational cost from iterative denoising, and (ii) limited ability to enforce constraints, as rewards bias generation toward desirable behaviors but do not guarantee constraint satisfaction.

\noindent\textbf{Conditional flow matching.}
Conditional flow matching (CFM) \cite{LipmanChenEtAl2023} has recently emerged as a promising alternative to diffusion models.
By learning a continuous velocity field that transports noise samples to the data distribution, CFM can achieve comparable generation quality with fewer function evaluations at inference time.
This efficiency makes CFM particularly attractive for real-time robot planning \cite{ChisariHeppertEtAl2024,YeGombolay2024}. 
Classifier-free guidance, originally developed for diffusion models \cite{HoSalimans2021}, has been extended to CFM \cite{ZhengLeEtAl2023}, but typically requires labeled conditioning signals that may not be available in robotics settings.
An alternative is reward-based guidance, which can bias generation at inference time without additional labels \cite{JannerDuEtAl2022}.
Our work builds on this direction by developing a reward-guided CFM framework that uses control barrier function-based rewards to generate safety-aware trajectory priors for robot navigation.

\section{Preliminaries: Conditional Flow Matching} \label{sec:preliminaries}

Conditional flow matching (CFM) \cite{LipmanChenEtAl2023} is a simulation-free method for training continuous normalizing flows \cite{ChenRubanovaEtAl2018} that transport a simple source distribution $p$ into a target data distribution $q$.
Let $\mathbf{z}_1 \in \mathbb{R}^{d_z}$ denote a data sample drawn from the conditional target distribution $q(\mathbf{z}_1\mid\mathbf{c})$, where $\mathbf{c} \in \mathbb{R}^{d_c}$ is a conditioning variable, and let $\mathbf{z}_0 \sim p(\mathbf{z}_0)$ be a sample from the source distribution, which we take to be $p(\mathbf{z}_0)=\mathcal{N}(0,I)$.
CFM learns a velocity field $v_\theta: \mathbb{R}^{d_z} \times \mathbb{R}^{d_c} \times [0,1] \rightarrow \mathbb{R}^{d_z}$
that defines an ordinary differential equation (ODE) $\frac{d\mathbf{z}_\tau}{d\tau}=v_\theta(\mathbf{z}_\tau,\mathbf{c},\tau)$. 
To avoid confusion with physical time, we use $\tau\in[0,1]$ to denote the flow time.
Solving this ODE from $\tau = 0$ to $\tau = 1$ transports a sample from the source distribution to a sample from the target distribution $q(\mathbf{z}_1\mid\mathbf{c})$.

A key idea in CFM is to define a probability path $p_\tau(\mathbf{z}\mid\mathbf{c})$ that smoothly interpolates between the source distribution at $\tau = 0$ and the target distribution at $\tau = 1$:
\begin{equation}
\begin{aligned}
    &p_\tau(\mathbf{z} \mid \mathbf{c})=\int p_{\tau\mid1}(\mathbf{z} \mid \mathbf{z}_1)q(\mathbf{z}_1 \mid \mathbf{c})d\mathbf{z}_1,\\ 
    &\mathrm{where}\ p_{\tau \mid 1}(\mathbf{z} \mid \mathbf{z}_1)=\mathcal{N}(\tau\mathbf{z}_1, (1-\tau)^2I).\\
\end{aligned}
\label{eq:probability path}
\end{equation}
During training, we sample a flow time $\tau\sim U(0,1)$, data points $\mathbf{z}_1\sim q(\mathbf{z}_1 \mid \mathbf{c})$, and noise $\mathbf{z}_0\sim p(\mathbf{z}_0)$.
Under the conditional probability path defined by $\mathbf{z}_1$, the intermediate point is given by the linear interpolation $\mathbf{z}_\tau=(1-\tau)\mathbf{z}_0+\tau \mathbf{z}_1$ whose time derivative is $\dot{\mathbf{z}}_\tau=\mathbf{z}_1-\mathbf{z}_0$.
Accordingly, the velocity field is parameterized by a neural network  $v_\theta$ and trained using the CFM loss
\begin{align}\label{eq:cfm_loss}
    L_\mathrm{CFM}=\mathbb{E}_{\tau,q(\mathbf{z}_1,\mathbf{c}),p(z_0)}[\|v_\theta(\mathbf{z}_\tau,\mathbf{c},\tau) - (\mathbf{z}_1-\mathbf{z}_0)\|_2^2].
\end{align}
During inference, we sample $\mathbf{z}_0\sim p(\mathbf{z}_0)$ and solve the ODE with the learned velocity field $v_\theta$ from $\tau=0$ to $\tau=1$ to generate a sample from the conditional distribution $q(\mathbf{z}_1 \mid \mathbf{c})$.

\section{Problem Formulation} \label{sec:problem}
This work addresses the problem of generating safe and efficient trajectories for a robot navigating through dynamic environments populated by humans. 
We formulate this as a finite-horizon optimal control problem, where the goal is to compute a sequence of control inputs that minimizes a task-dependent objective while satisfying system dynamics and safety requirements.

Let a robot state trajectory be denoted as $\mathbf{x}=[\mathbf{x}_0,\ \ldots,\ \mathbf{x}_{T}]$ where $\mathbf{x}_t \in \mathcal{D}\subseteq\mathbb{R}^{d_x}$ and control inputs as $\mathbf{u}=[\mathbf{u}_0,\ \ldots,\ \mathbf{u}_{T-1}]$ where $\mathbf{u}_t \in \mathcal{U} \subseteq \mathbb{R}^{d_u}$, where $T$ represents the planning horizon.
The robot evolves according to the discrete-time dynamics $\mathbf{x}_{t+1}=f(\mathbf{x}_t, \mathbf{u}_t)$ where $f$ denotes the robot dynamics model.
The environment contains $N$ dynamic human agents, and the state of the $i$-th human at time $t$ is denoted by $\mathbf{x}_{h_i,t}$.

Given the current robot state and the surrounding human states, the planning problem is to find a control sequence $\mathbf{u}$ that minimizes a cost function $J(\mathbf{x}, \mathbf{u})$. 
In our setting, this objective primarily captures progress toward the goal while penalizing unsafe proximity to nearby humans. 
A key requirement is safety, which we formulate as maintaining a minimum separation distance $r_{\mathrm{safe}}$ from every human at all times:
\begin{align}
\|\mathbf{p}_t - \mathbf{p}_{h_i,t}\|2 \ge r_{\mathrm{safe}},
\quad \forall i \in \{1,\ldots,N\},\ \forall t \in \{0,\ldots,T\},
\end{align}
where $\mathbf{p}_t$ and $\mathbf{p}_{h_i, t}$ are the positions of the robot and the $i$-th human at time $t$, respectively.
The resulting planning problem is

\vspace{-3mm}
{\small
\begin{equation}
\begin{aligned}
\min_{\mathbf{u}} & \quad J(\mathbf{x}, \mathbf{u}) \\
\text{s.t.} & \quad \mathbf{x}_{t+1}=f(\mathbf{x}_t, \mathbf{u}_t), \: \mathbf{x}_0 = \mathbf{x}_{\text{init}}, \:  t=1,\ldots,T-1\\
& \quad \|\mathbf{p}_t-\mathbf{p}_{h_i, t}\|_2 \geq r_{\text{safe}}, \:  i =1,\ldots,N, t=0,\ldots,T \label{eq:safe_const}\\
& \quad \mathbf{u}_t \in \mathcal{U}, \: t=0,\ldots,T-1.
\end{aligned}
\end{equation}
}
The main challenge is solving this problem online in dynamic, uncertain, and multimodal environments where future human motions are not known in advance. 
Our framework addresses this challenge by combining a generative model that produces diverse trajectory priors with an optimization-based controller that refines them online to improve safety and goal-reaching performance.

\section{Guided Conditional Flow Matching for Trajectory Optimization} \label{sec:guided flow matching}

In this section, we describe how conditional flow matching (CFM) is used to generate informed control-sequence candidates for sampling-based MPC. 
We first train a conditional CFM model over control sequences, and then introduce reward guidance at inference time to steer generation toward desirable behaviors without retraining.

\subsection{CFM model training}
In this work, the CFM model transforms random noise into a control sequence $\mathbf{u}$. 
Accordingly, the flow variable $\mathbf{z}$ introduced in Sec.~\ref{sec:preliminaries} corresponds to $\mathbf{u}$. 
The target distribution is defined over control sequences from the training dataset and is conditioned on $\mathbf{c}$, which includes the robot's current state, goal position, and previously executed controls.

\noindent\textbf{Conditioning on past controls.} 
To encourage smooth and temporally coherent motion, we include a history of previously executed controls in the conditioning variable $\mathbf{c}$. 
Rather than conditioning only on the current environment and goal, we also provide recent control history and train the model to complete the remaining sequence accordingly. 
The history length induces a trade-off: longer histories improve continuity with past actions, but may reduce responsiveness to rapid environmental changes. 
This conditioning mechanism helps the model generate control sequences that remain consistent across replanning steps.

Training the CFM model with this conditioning yields a velocity field $v_\theta$. 
We next show how this learned field can be guided online using a user-defined reward function, enabling flexible behavior adaptation at inference time.

\subsection{CFM guidance with reward functions} \label{subsec:guidance}
To bias CFM sampling toward desirable behaviors, we incorporate reward-based guidance at inference time. 
Inspired by reward-guided diffusion methods \cite{JannerDuEtAl2022}, we adapt this idea to flow matching by modifying the learned velocity field using gradients of a differentiable reward evaluated on the predicted terminal trajectory.

\begin{proposition}[Reward-guided CFM Velocity Field]
Given a base velocity field $v_\theta(\mathbf{z}_\tau,\mathbf{c},\tau)$ and a differentiable reward function $R(\mathbf{z}_1)$, the \textit{guided} velocity field $\tilde{v}_\theta(\mathbf{z}_\tau,\mathbf{c},\tau)$ is given by
\begin{align}\label{eq:CFM guidance}
    \tilde{v}_\theta(\mathbf{z}_\tau,\mathbf{c},\tau) = v_\theta(\mathbf{z}_\tau,\mathbf{c},\tau) + \lambda_\text{guide}\nabla_{\mathbf{z}_1} R(\mathbf{z}_1),
\end{align}
where $\lambda_\text{guide}$ controls the guidance strength.
\end{proposition}
\begin{proof}
Following the control-as-inference formulation in \cite{JannerDuEtAl2022,Levine2018}, let $\mathbf{y}$ be a binary random variable indicating optimality, where $\mathbf{y}=1$ denotes an optimal sample. 
We define $p_\tau(\mathbf{y}=1 \mid \mathbf{z}_1) = \exp \left(R(\mathbf{z_1})\right)$ where $R(\mathbf{z_1})$ is a reward function on the terminal sample $\mathbf{z}_1$.
For notational simplicity, we omit ``$=1$'' in the remainder of the derivation.
Using \cite[Lem.~1]{ZhengLeEtAl2023} on \eqref{eq:probability path}, we can apply Bayes' rule to the conditional velocity field $v_\tau(\mathbf{z}_\tau\mid\mathbf{y})$ and then group terms depending only on $\mathbf{z}_\tau$,

\vspace{-3mm}
{\small
\begin{align*}
    v_\tau(\mathbf{z}_\tau\mid\mathbf{y}) &= \frac{1}{\tau}\mathbf{z}_\tau+\frac{1-\tau}{\tau}\nabla_{\mathbf{z}_\tau}\log p_\tau(\mathbf{z}_\tau\mid\mathbf{y}) \: \text{(\cite[Lem.~1]{ZhengLeEtAl2023})}\\
    &= \frac{1}{\tau}\mathbf{z}_\tau+\frac{1-\tau}{\tau}\nabla_{\mathbf{z}_\tau}\log \left( \frac{p_\tau(\mathbf{y} \mid \mathbf{z}_\tau)p_\tau(\mathbf{z}_\tau)}{p_\tau(\mathbf{y})} \right)\\
    &=\frac{1}{\tau}\mathbf{z}_\tau+\frac{1-\tau}{\tau}\left(\nabla_{\mathbf{z}_\tau}\log p_\tau(\mathbf{z}_\tau)+\nabla_{\mathbf{z}_\tau}\log p_\tau(\mathbf{y}\mid\mathbf{z}_\tau)\right)\\
    &=v_\tau(\mathbf{z}_\tau)+\frac{1-\tau}{\tau}\nabla_{\mathbf{z}_\tau}\log p_\tau(\mathbf{y}\mid\mathbf{z}_\tau),  \: \text{(\cite[Lem.~1]{ZhengLeEtAl2023})}
\end{align*}
}
where $v_\tau(\mathbf{z}_\tau)$ denotes the base velocity field.
Given the relationship between intermediate and predicted states $\tau\mathbf{z}_1=\mathbf{z}_\tau-(1-\tau)\mathbf{z}_0$, we use the chain rule to express the gradient term,
\begin{align*}
    v_\tau(\mathbf{z}_\tau\mid\mathbf{y}) &=v_\tau(\mathbf{z}_\tau)+\frac{1-\tau}{\tau}(\nabla_{\mathbf{z}_\tau}\mathbf{z}_1)^T\nabla_{\mathbf{z}_1}\log p_\tau(\mathbf{y}\mid\mathbf{z}_1)\\
    &= v_\tau(\mathbf{z}_\tau)+\frac{1-\tau}{\tau^2}\nabla_{\mathbf{z}_1}\log p_\tau(\mathbf{y}\mid\mathbf{z}_1).
\end{align*}
Then following from \cite{JannerDuEtAl2022} where $p_\tau(\mathbf{y} \mid \mathbf{z}_1) = \exp \left(R(\mathbf{z_1})\right)$, we have $\log p_\tau(\mathbf{y}\mid\mathbf{z}_1) = R(\mathbf{z}_1)$.
Finally, we absorb the factor $\frac{1-\tau}{\tau^2}$ into the guidance coefficient $\lambda_\mathrm{guide}$, yielding \eqref{eq:CFM guidance}.
\end{proof}

\noindent\textbf{Reward guidance steps.} 
To steer this process with a reward function, we introduce a multi-step iterative refinement, as summarized in Algorithm~\ref{alg:guidedCFM}. 
This addresses the challenge that the guidance term $R(\mathbf{z}_1)$ depends on the final trajectory $\mathbf{z}_1$, which is not directly available at intermediate times $\tau_i$.

Given a user-defined ODE schedule $\{\tau_0,\ldots,\tau_N\}$, at each step $\tau_i$, we first form a temporary estimate of the final trajectory $\mathbf{z}_1$ by integrating the current \textit{base} velocity field $v_\theta(\mathbf{z}_{\tau_i}, \mathbf{c}, \tau_i)$ from $\tau=\tau_i$ to $\tau=1$ (Line \ref{algstep:get z1}).
We then evaluate the reward gradient with respect to this predicted terminal sample (Line \ref{algstep:reward gradient}).
This gradient is added to the base velocity field, yielding the \textit{guided velocity field}, as shown in \eqref{eq:CFM guidance} (Line \ref{algstep:guided velocity field}). 
Finally, we use this new guided velocity field to take a single integration step from $\tau_i$ to the next point $\tau_{i+1}$.

\noindent\textbf{Guided CFM vs guided Diffusion.} 
A key advantage of CFM over diffusion models is sampling efficiency. 
Diffusion models typically require many denoising steps, whereas CFM generates samples by integrating a learned velocity field over a much smaller number of function evaluations. 
Moreover, unlike guided diffusion methods that often evaluate rewards on noisy intermediate samples \cite{MizutaLeung2024}, our approach evaluates rewards on predicted noise-free trajectories. 
Although our procedure remains iterative, it retains a substantial efficiency advantage and is well-suited to real-time replanning without retraining the base CFM.

\subsection{Reward function} 
The reward function is user-defined and can, in principle, combine multiple behavioral objectives. 
In this work, we focus on two reward components relevant to social navigation: safety and goal reaching. 

For safety, we employ control barrier functions (CBFs) \cite{AmesCooganEtAl2019} to define a reward over states and controls. 
Specifically, we define the safety reward $r_\mathrm{safe}(\mathbf{x}_t,\mathbf{u}_t)$ as the degree of violation of the CBF condition:
\begin{align}
    &r_\mathrm{safe}(\mathbf{x}_t, \mathbf{u}_t) = \gamma_t\cdot\min\{0,\,\dot{h}(\mathbf{x}_t) + \alpha(h(\mathbf{x}_t))\},\\
    &h(\mathbf{x}_t) = \|\mathbf{p}_t - \mathbf{p}_\mathrm{h_i,t}\|^2 - r^2,\notag
\end{align}
where $\dot{h}(\mathbf{x}_t)$ denotes the time derivative of $h(\mathbf{x}_t)$, $\alpha$ is an extended class-$\mathcal{K}$ function that bounds the rate at which the system approaches the unsafe set, and $\gamma_t$ is a markup term that encourages proactive collision avoidance \cite{GeldenbottLeung2024}.
Although the CBF reward depends on both states and controls, our CFM model generates only a control sequence. 
We therefore roll out the dynamics $\mathbf{x}_{t+1}=f(\mathbf{x}_t,\mathbf{u}_t)$ under the generated controls to obtain the corresponding state trajectory, and use automatic differentiation to compute reward gradients with respect to the controls.

In addition to safety, we include a goal-reaching reward $r_\mathrm{goal}$ to encourage progress toward the target. 
The total reward over the planning horizon is
\begin{align}
R(\mathbf{u}) = \sum_{t=0}^{T-1}  w_\mathrm{safe} r_\mathrm{safe}(\mathbf{x}_t, \mathbf{u}_t) + w_\mathrm{goal}  r_\mathrm{goal}(\mathbf{x}_t), 
\end{align}
where $w_\mathrm{safe}$ and $w_\mathrm{goal}$ are scalar weights. 
While our experiments focus on these two components, the same framework can incorporate additional differentiable reward terms. 
The specific reward definitions and weights used in our experiments are provided in Sec.~\ref{sec:evaluation}.

\begin{algorithm}[t]
\caption{Reward-Guided CFM Trajectory Generation}
\label{alg:guidedCFM}
\begin{algorithmic}[1]
\Require Conditioning variable $\mathbf{c}$, guidance strength $\lambda_\text{guide}$, ODE schedule $\{\tau_0,\ldots,\tau_N\}$
\State Initialize: $\mathbf{z}_0 \sim \mathcal{N}(0,I)$
\For{$i=0$ \textbf{to} $N-1$}
    \State Compute base velocity field: $v_\theta(\mathbf{z}_{\tau_i}, \mathbf{c}, \tau_i)$ \label{algstep:velocity field}
    \State Integrate $\mathbf{z}_{\tau_i}$ from $\tau=\tau_i$ to $\tau=1$ to obtain $\mathbf{z}_{1}$ \label{algstep:get z1}
    \State Compute gradient $g = \nabla_{\mathbf{z}_1} R(\mathbf{z}_1)$  \label{algstep:reward gradient}
    \State $\tilde{v}_\theta \gets v_\theta(\mathbf{z}_{\tau_i}, \mathbf{c}, \tau_i) + \lambda_\text{guide}\cdot g$ \hfill (See \eqref{eq:CFM guidance}) \label{algstep:guided velocity field}
    \State $\mathbf{z}_{\tau_{i+1}}\leftarrow\mathbf{z}_{\tau_i}+\tilde{v}_\theta(\tau_{i+1}-\tau_i)$ \label{algstep:integrate step}
\EndFor
\State \Return $\mathbf{z}_{\tau_N}$
\end{algorithmic}
\end{algorithm}

\section{Bridging Flow Matching with Trajectory Optimization}

\begin{figure}[t]
  \centering
  \captionsetup{width=\linewidth}
  \includegraphics[width=\linewidth]{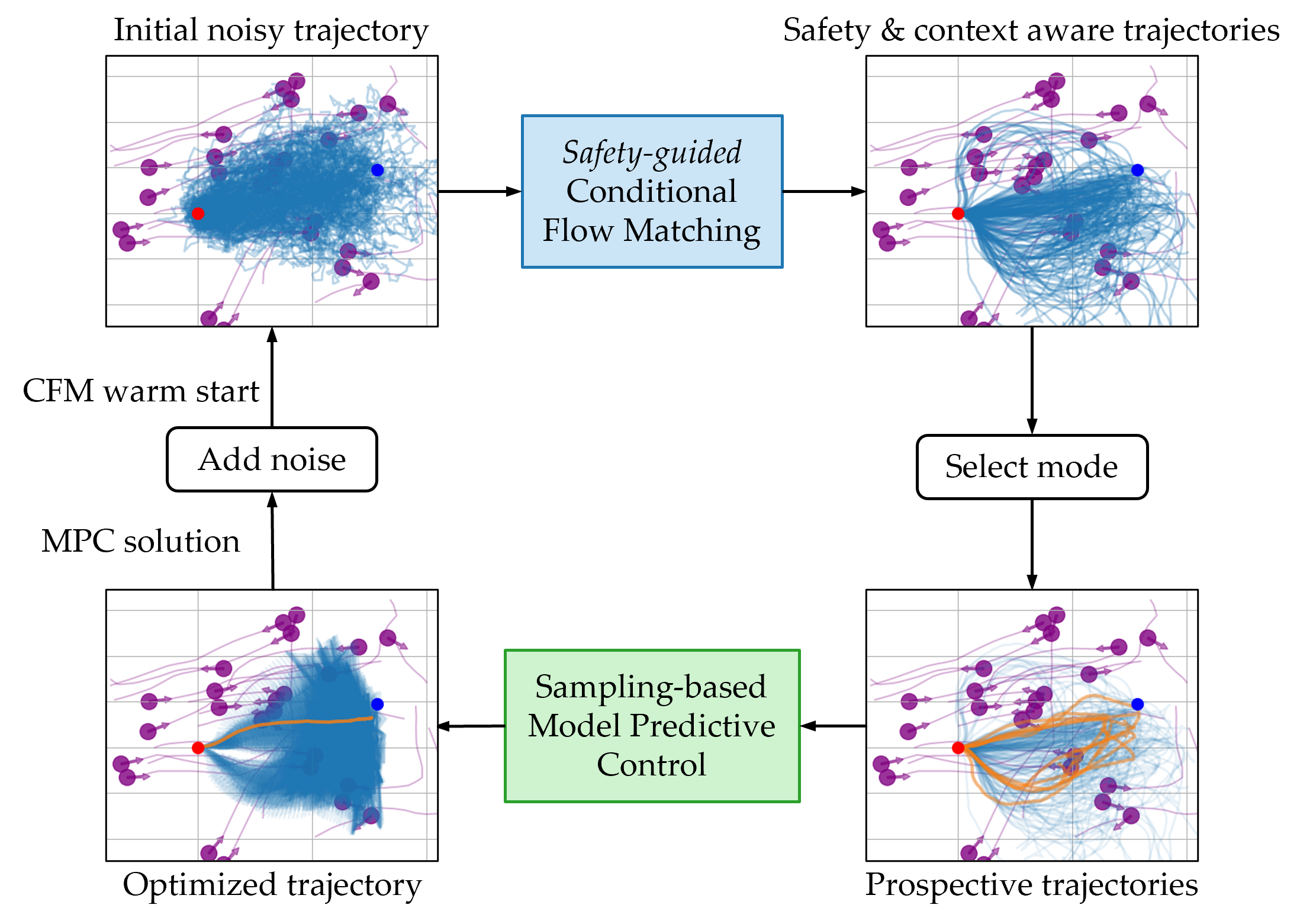}
  \caption{A reward-guided conditional flow matching (CFM) model generates diverse trajectory priors for model predictive control (MPC), whose optimized solution warm-starts the next CFM sampling step.}
  \label{fig:proposed_framework}
  \vspace{-5mm}
\end{figure}

We integrate guided CFM with sampling-based MPC in a bidirectional generation-refinement loop. 
Guided CFM generates trajectory priors for MPC, and the optimized MPC trajectory warm-starts the next CFM sampling step. 
In this paper, we instantiate the MPC module with model predictive path integral (MPPI) control \cite{WilliamsDrewsEtAl2016} and refer to the resulting method as \textbf{CFM-MPPI}, although the same framework can be applied to other sampling-based MPC methods. 
Fig.~\ref{fig:proposed_framework} illustrates this information flow between the CFM and MPC modules.

\subsection{Benefits of integrating CFM with MPPI}
Although guided CFM biases sampling toward high-reward regions, the generated trajectories may still be suboptimal, unsafe, or dynamically inconsistent with the deployment robot. 
An additional optimization stage is therefore useful for refining these trajectories online with respect to safety, task performance, and system dynamics.

Using multiple CFM-generated candidates is also advantageous in nonconvex planning problems. 
Rather than relying on a single learned trajectory, the optimizer can refine several promising candidates, improving robustness to local minima and enabling better search over distinct behavioral modes.

A further benefit is adaptability across robot embodiments. 
The control patterns captured in the dataset may not exactly match the dynamics of the target robot, and directly executing the generated controls can therefore be suboptimal. 
By refining the sampled trajectories under the robot's actual dynamics, the proposed framework better aligns learned priors with the deployment system.

\subsection{Mode selection: Handling multimodal priors from CFM}
Unlike standard MPPI, which samples Gaussian perturbations around a single reference trajectory, our approach starts from multiple trajectory candidates $\{\mathbf{u}_k\}_{k=1}^K$ generated by guided CFM. 
These candidates serve as learned trajectory priors that capture multiple plausible modes of behavior.

This creates a mismatch with the standard MPPI update, which is designed for a unimodal local sampling distribution. 
If trajectories from distinct modes are combined through a single cost-weighted average, the resulting trajectory can lie in an undesirable or infeasible region between modes, degrading optimization performance.

To address this, our approach first evaluates all trajectory candidates generated by the CFM and selects the top $K^*$ trajectories with the lowest cost.
Each selected trajectory is then used as a reference trajectory for an independent MPPI refinement run, executed in parallel. 
The lowest-cost refined trajectory is chosen for execution. 
By refining promising modes separately, the proposed approach preserves multimodal structure and avoids collapsing distinct modes into a suboptimal intermediate solution.

\subsection{Warmstarting CFM with MPC solutions}
To close the feedback loop from optimization back to generation, we warm-start CFM using the optimized control sequence returned by MPC. 
Instead of initializing sampling from the standard Gaussian prior, we initialize the flow at an intermediate time $\tau \in (0,1)$ using a noised version of the previous MPC solution.

Specifically, given the previous optimal control sequence $\mathbf{u}$, we apply the forward noising process to obtain an intermediate latent state $\mathbf{z}_\tau$. 
The flow is then integrated from $\mathbf{z}_\tau$ to $\mathbf{z}_1$, so that the next sample is generated by refining the previous MPC solution rather than starting from scratch. 
This promotes temporal consistency across replanning steps and can reduce the effort required to generate high-quality trajectories online.

The amount of injected noise determines the trade-off between consistency and exploration. 
With small noise, the generated trajectory remains close to the previous optimum, yielding smoother and more predictable behavior. 
With larger noise, the generator explores a wider set of candidate trajectories, which can be beneficial when the environment changes significantly.

\section{Experiments and Discussion} \label{sec:evaluation}

We evaluate the proposed method on social navigation tasks in dynamic human environments using robots with unicycle and double-integrator dynamics.

\subsection{Experimental setup} \label{subsec:settings}

\noindent\textbf{Train and test data.} 
We train the CFM model on the ETH pedestrian dataset \cite{PellegriniEssEtAl2009}, which contains $276{,}874$ trajectories between 1 and 8 seconds long.
The learned trajectories are interpreted under single-integrator dynamics and mapped to control inputs for both unicycle and double-integrator robots. 
For the unicycle model, this mapping uses the robot orientation and a control-point offset: the linear velocity is obtained by projecting the input velocity onto the robot heading, while the angular velocity is chosen so that the control point follows the path. 
For the double-integrator model, the target velocities generated by the CFM are tracked directly using acceleration inputs.

For evaluation, we use the UCY pedestrian dataset \cite{LernerChrysanthouEtAl2007}, the SDD dataset \cite{RobicquetSadeghianEtAl2016}, which includes faster-moving agents such as bicyclists, and a simulated crowd environment with 20 agents modeled by social force models (SFMs) \cite{HelbingMolnar1995}. 
We evaluate 300 test scenarios for each environment and each dynamics model. 
At each planning step, we assume the robot has access to the exact positions and velocities of surrounding agents, and predicts their future trajectories using a constant-velocity model. 
For the SFM simulations, we additionally test robustness to observation noise by perturbing obstacle positions with Gaussian noise of standard deviation $0.05$.

\noindent\textbf{CFM model details.} 
We use a transformer backbone \cite{VaswaniShazeerEtAl2017} for the CFM model.
For the initial reward-guided trajectory generation, we use a 9-step ODE schedule: $[0.0, 0.5, 0.8, 0.85, 0.9, 0.92, 0.94, 0.96, 0.98, 1.00]$.
At subsequent planning steps, generation is warm-started from $\tau=0.8$ using the previous MPPI solution, reducing the process to the final 7 steps.
We set the guidance strength to $\lambda_\text{guide}=1.0$, the goal weight to $w_\mathrm{goal}=0.1$, the markup factor to $1.01$, and the collision radius to $r_\mathrm{safe}=0.5$m.
To generate diverse trajectories ranging from aggressive to conservative behaviors, we use five safety weights $w_\mathrm{safe}=[0.1, 0.3, 0.5, 0.7, 0.9]$, and generate 40 trajectories for each weight, resulting in 200 trajectories in total.

\noindent\textbf{Guided CFM reward functions.} 
For CFM guidance, we use the reward defined in Sec.~\ref{subsec:guidance}, instantiated with a CBF-based safety reward and a terminal goal-reaching reward. 
The goal reward encourages the robot to reach its designated target $\mathbf{x}_g$ by penalizing the squared Euclidean distance to the goal.

\noindent\textbf{Cost functions for MPPI.} 
The MPPI objective mirrors the reward structure used for CFM guidance and includes both safety and goal-reaching terms. 
For safety, rather than using a CBF-based term, we define an exponential proximity cost:
\begin{align}
    c_\text{safe}(\mathbf{p}_t) = \exp(-\beta( \|\mathbf{p}_t - \mathbf{p}_\mathrm{h_i,t}\| - r)),
\end{align}
where $\beta$ controls the steepness of the potential field. The goal-reaching term encourages progress toward the target.

\begin{figure}[t]
\centering
\begin{minipage}[t]{0.32\linewidth}
  \centering
  \includegraphics[width=\linewidth]{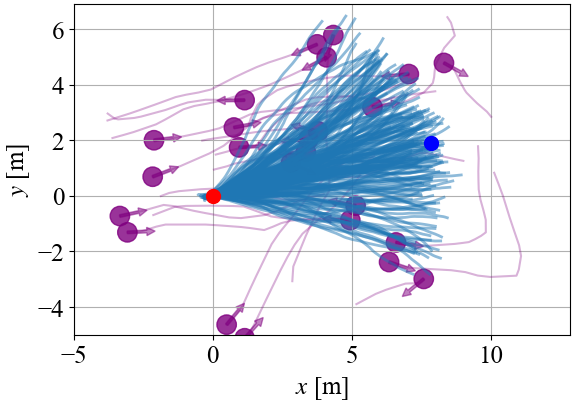}
  \subcaption{Gaussian.}
\end{minipage}
\begin{minipage}[t]{0.32\linewidth}
  \centering
  \includegraphics[width=\linewidth]{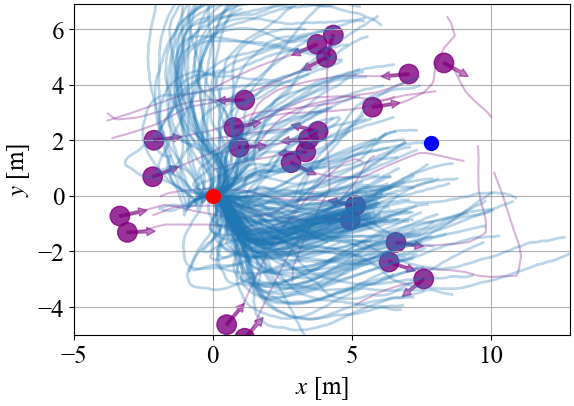}
  \subcaption{Diffusion.}
\end{minipage}
\begin{minipage}[t]{0.32\linewidth}
  \centering
  \includegraphics[width=\linewidth]{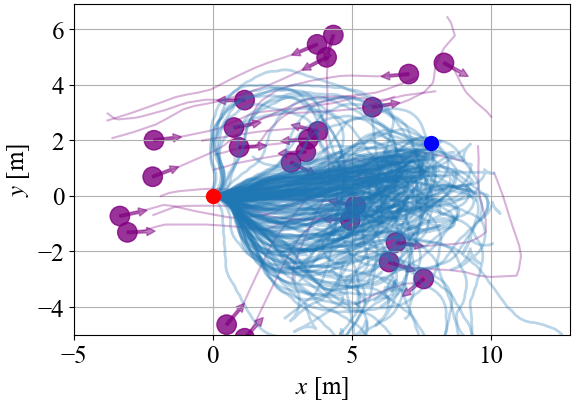}
  \subcaption{CFM.}
\end{minipage}
\captionsetup{width=\linewidth}
\caption{Trajectories generated from different sampling distributions. Pedestrians and their future paths are shown in purple. The robot is shown as a red dot and the goal as a blue dot.}
\label{fig:prior}
\end{figure}

\noindent\textbf{Comparison methods.}
We compare against the following baselines.
\textbf{\textit{MPPI}}: standard MPPI with a Gaussian sampling distribution.
\textbf{\textit{CFM}}: reward-guided CFM without MPPI; the lowest-cost sampled trajectory is selected for execution.
\textbf{\textit{Diff-MPPI}}: a guided diffusion model used to generate trajectory priors for MPPI.
\textbf{\textit{CFM-MPPI}}: a guided CFM model used to generate trajectory priors for MPPI.
\textbf{\textit{Diff-MPPI$^*$, CFM-MPPI$^*$}}: variants using mode-selective MPPI; unmarked versions apply MPPI directly to all generated samples.

Our MPPI implementation follows the algorithm in \cite{WilliamsDrewsEtAl2016}, with temperature parameter $\lambda=0.1$, and perturbation covariance $\sigma=[0.3,0.6]$ for the unicycle model and $\sigma=[0.4,0.4]$ for the double integrator model.
For the diffusion baseline, we use the same transformer backbone as CFM.
To keep the diffusion baseline within a comparable real-time budget (approximately 10\,Hz planning), we use DDIM inference. 
Although the model is trained with 1000 denoising steps, we use 50 denoising steps at test time and apply guidance only during the final 10 steps.
During inference, we use the same reward structure as for CFM, but with safety weights $w_\mathrm{safe}=[0.1, 0.2, 0.3, 0.4, 0.5]$, since diffusion evaluates reward gradients on noisy trajectories and otherwise tends to become overly conservative.

\noindent\textbf{Evaluation metrics.}
We report the following metrics:
\textbf{\textit{Collision}}: percentage of simulations that violate the collision radius.
\textbf{\textit{Reaching}}: Euclidean distance between the final robot position and the goal.
\textbf{\textit{Time}}: computation time required to compute the control input at each planning step.

\subsection{Qualitative results}
We first present qualitative results to illustrate the sampling behavior of the proposed method and the effect of guidance and warm-starting.

\noindent\textbf{Trajectory quality.}
Fig.~\ref{fig:prior} shows the sampling distribution produced by each method. 
CFM generates diverse trajectories concentrated in lower-density regions of the scene, whereas diffusion produces more spatially scattered candidates. 
In contrast, the Gaussian sampling distribution used by standard MPPI is not conditioned on the environment, leading to trajectories that often pass through denser areas of pedestrian traffic.

\begin{figure}[t]
\centering
\begin{minipage}[t]{0.32\linewidth}
  \centering
  \includegraphics[width=\linewidth]{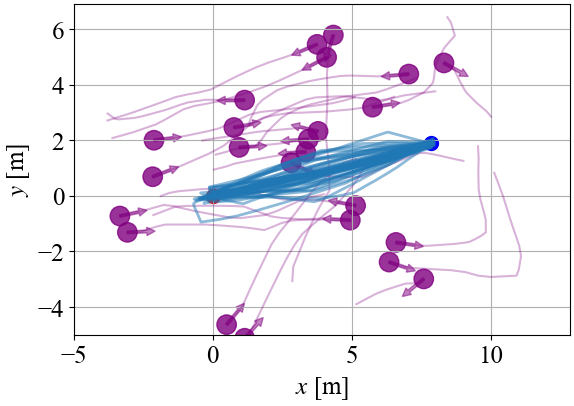}
  \subcaption{\centering $w_\text{safe}=0$.}
\end{minipage}
\begin{minipage}[t]{0.32\linewidth}
  \centering
  \includegraphics[width=\linewidth]{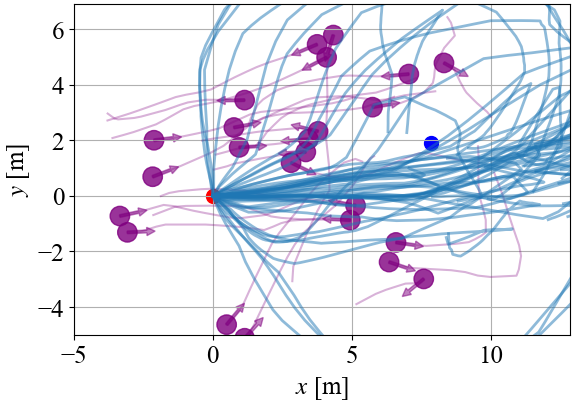}
  \subcaption{$w_\text{safe}=0.5$.}
\end{minipage}
\begin{minipage}[t]{0.32\linewidth}
  \centering
  \includegraphics[width=\linewidth]{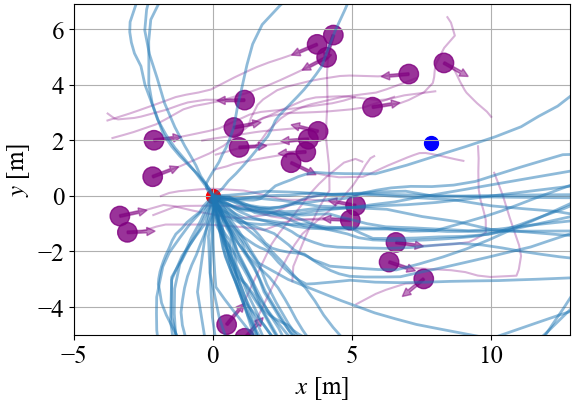}
  \subcaption{$w_\text{safe}=0.9$.}
\end{minipage}
\captionsetup{width=\linewidth}
\caption{Impact of CBF guidance weight $w_\text{safe}$ on trajectory generation using reward-guided CFM. Pedestrians and their future paths are shown in purple. The robot is shown as a red dot and the goal as a blue dot.}
\label{fig:guidance}
\vspace{-5mm}
\end{figure}

\begin{figure}[t]
\centering
\begin{minipage}[t]{0.32\linewidth}
  \centering
  \includegraphics[width=\linewidth]{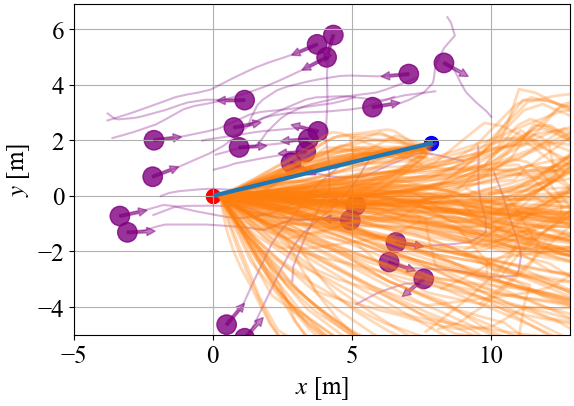}
  \subcaption{$\tau=0.0$.}
  \vspace{-3pt}
\end{minipage}
\begin{minipage}[t]{0.32\linewidth}
  \centering
  \includegraphics[width=\linewidth]{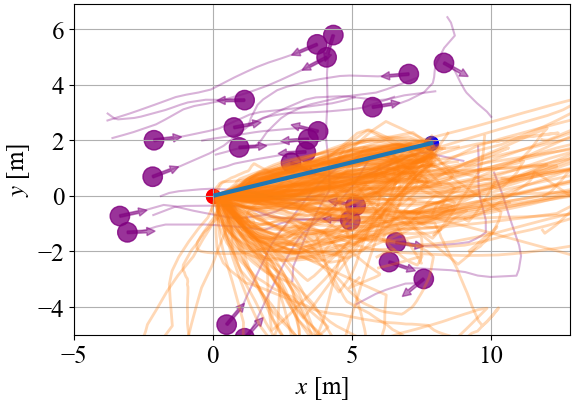}
  \subcaption{$\tau=0.5$.}
  \vspace{-3pt}
\end{minipage}
\begin{minipage}[t]{0.32\linewidth}
  \centering
  \includegraphics[width=\linewidth]{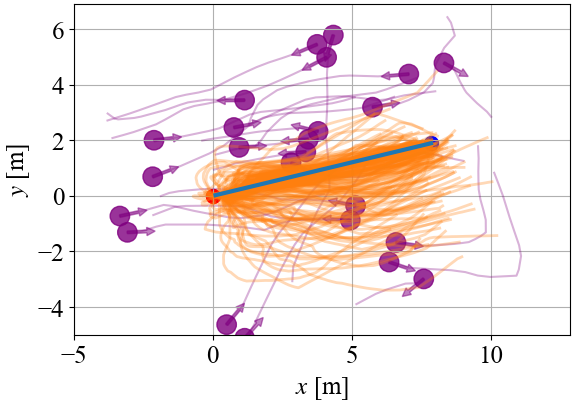}
  \subcaption{$\tau=0.8$.}
  \vspace{-3pt}
\end{minipage}
\captionsetup{width=\linewidth}
\caption{CFM trajectories generated using warm-starts of varying strengths $\tau$. The previous optimal solution is shown in blue. Pedestrians and their future paths are shown in purple.  The robot is shown as a red dot and the goal as a blue dot.}
\vspace{-5pt}
\label{fig:warm_start}
\end{figure}

\begin{figure*}[t]
  \centering
  \includegraphics[width=\linewidth]{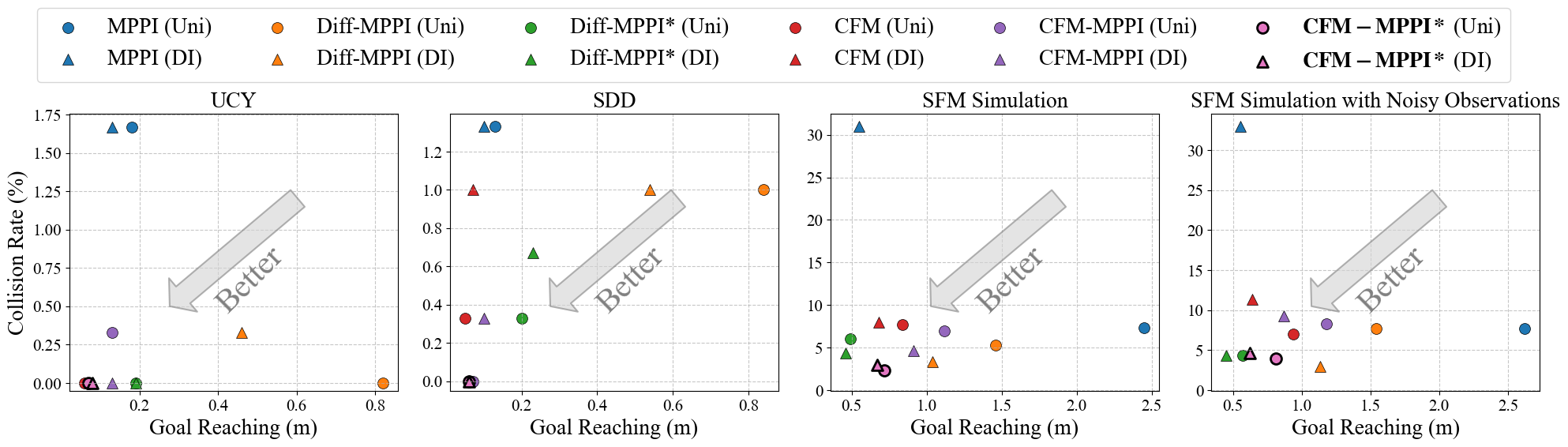}
  \captionsetup{width=\linewidth}
  \caption{
  Quantitative comparison of goal reaching versus collision rate across datasets and simulated environments. Colors indicate methods, while marker shapes denote the robot dynamics model.
  }
  \label{fig:safe_goal}
  \vspace{-5mm}
\end{figure*}

\begin{figure}[t]
  \centering
  \includegraphics[width=\linewidth]{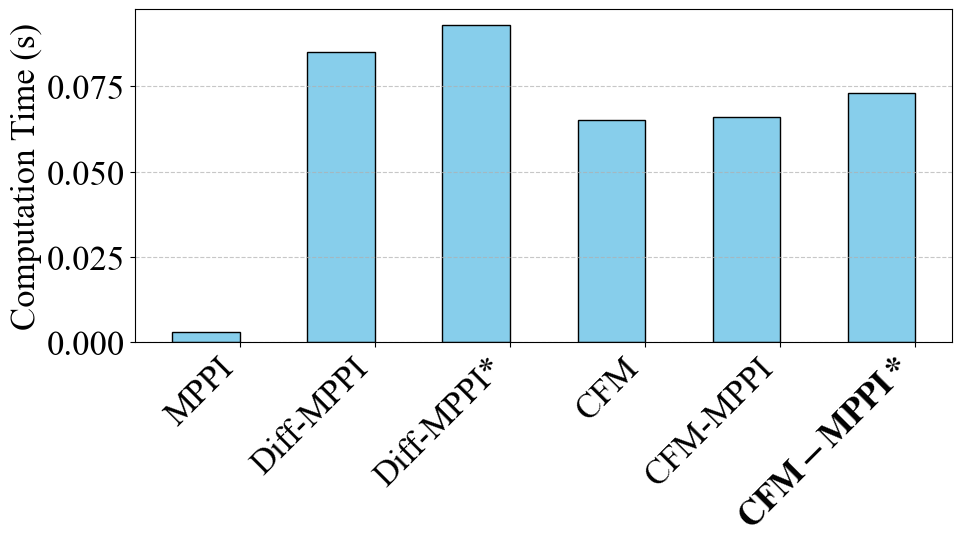}
  \captionsetup{width=\linewidth}
  \caption{
  Comparison of computation time across methods on the UCY dataset.
  }
  \label{fig:time}
  \vspace{-5mm}
\end{figure}

\noindent\textbf{CFM guidance.}
Fig.~\ref{fig:guidance} illustrates how varying the CBF guidance weight $w_\mathrm{safe}$ changes the conservativeness of the generated trajectories. 
Smaller values of $w_\mathrm{safe}$ produce trajectories that move more directly toward the goal, whereas larger values encourage stronger obstacle avoidance. 
The generated trajectories are also clearly \textit{multimodal}, clustering around different feasible gaps between nearby humans.

\noindent\textbf{CFM warm-starting.}
Fig.~\ref{fig:warm_start} illustrates the effect of warm-starting CFM. 
As less noise is injected into the previous optimal solution (equivalently, as $\tau$ increases), the generated trajectories remain closer to the previous trajectory. 
Stronger warm-starting is beneficial when the environment changes slowly and the previous solution remains informative, whereas weaker warm-starting encourages greater exploration when conditions change rapidly. 
In the current implementation, the warm-start strength is fixed; adapting it online is an interesting direction for future work.

\subsection{Quantitative results and key takeaways} \label{subsec:results}

Fig.~\ref{fig:safe_goal} summarizes the trade-off between goal reaching and collision rate across the pedestrian datasets, the SFM simulator, and the noisy-observation setting. 
Fig.~\ref{fig:time} compares computation time across methods.

\noindent \textbf{\textit{Takeaway 1}.}
\textit{CFM provides a favorable balance between safety and goal reaching compared with diffusion.}
CFM-based methods, especially CFM-MPPI$^*$, achieve strong safety performance while maintaining high goal-reaching accuracy.
Although Diff-MPPI occasionally attains slightly lower collision rates under noisy observations, this gain comes at the expense of goal-reaching performance. 
In such cases, the robot tends to slow down or take a longer path, preventing it from reaching the goal as closely within the allotted time. 
We hypothesize that diffusion guidance is less effective in balancing these objectives because reward gradients are evaluated on noisy intermediate trajectories. 
Increasing the number of denoising steps can improve diffusion performance, but at additional computational cost.

\noindent\textbf{\textit{Takeaway 2}.}
\textit{Mode selection improves robustness.} 
In the SFM simulator, where agents are reactive and observations can be noisy, the mode-selective variants (denoted by $^*$) perform noticeably better than their standard counterparts.
For instance, compared with CFM-MPPI, CFM-MPPI$^*$ roughly halves the collision rate in both the nominal and noisy simulated settings, while also improving goal reaching.
This highlights the benefit of refining promising candidate modes rather than aggregating all generated trajectories directly. 
It also suggests that more sophisticated selection strategies, for example, those that explicitly account for both quality and diversity, may further improve robustness.

\noindent\textbf{\textit{Takeaway 3}.}
\textit{Learned sampling priors substantially outperform uninformed priors in safety and task performance.}
Aside from computation time, the uninformed MPPI baseline performs worst overall, especially in the reactive SFM environment. 
While this trend is not surprising, the first two takeaways show that both the quality of the prior and the way it is incorporated into optimization strongly affect robustness. 
Standard MPPI struggles in complex interaction scenarios, whereas both CFM- and diffusion-based priors achieve much lower collision rates. 
These results underscore the importance of informative generative priors for safe navigation.

\noindent\textbf{\textit{Takeaway 4}.}
\textit{CFM offers a strong balance between computational cost and performance.}
As shown in Fig.~\ref{fig:time}, a key advantage of CFM is computational efficiency, with control sequences generated in under 0.1 seconds.
For both CFM and diffusion, the main computational bottleneck is reward evaluation at each guidance step.
Because CFM requires fewer guidance steps than diffusion, it also requires fewer reward evaluations. 
When diffusion is constrained to match the inference speed of CFM, its safety performance deteriorates. 
Thus, CFM provides higher-quality trajectories under a significantly tighter real-time budget.

\section{Conclusion} \label{sec:conclusion}
We presented a unified planning framework that integrates conditional flow matching (CFM) with sampling-based model predictive control (MPC) for dynamic multi-agent environments. 
Our approach uses a \textit{reward-guided} CFM model to generate informed, multimodal trajectory priors for sampling-based MPC, while feeding optimized MPC solutions back to warm-start subsequent CFM generation.
To effectively exploit these multimodal priors, we further introduced a mode-selective MPPI strategy that identifies promising trajectory modes and refines them in parallel, avoiding the collapse of distinct paths into a suboptimal intermediate trajectory. 
Experiments in social navigation showed that the proposed framework achieves a favorable balance between safety, goal reaching, and real-time performance across a range of dynamic environments.

\bibliographystyle{IEEEtran}
\bibliography{main,ctrl_papers}

\end{document}